%% file: main.tex
\pdfoutput=1
\documentclass[letterpaper]{article} % DO NOT CHANGE THIS
\usepackage{aaai23}  % DO NOT CHANGE THIS
\usepackage{times}  % DO NOT CHANGE THIS
\usepackage{helvet}  % DO NOT CHANGE THIS
\usepackage{courier}  % DO NOT CHANGE THIS
\usepackage[hyphens]{url}  % DO NOT CHANGE THIS
\usepackage{graphicx} % DO NOT CHANGE THIS
\usepackage{natbib}  % DO NOT CHANGE THIS AND DO NOT ADD ANY OPTIONS TO IT
\usepackage{caption} % DO NOT CHANGE THIS AND DO NOT ADD ANY OPTIONS TO IT
\usepackage{algorithm}
\usepackage{algorithmic}

\usepackage{bm}
\usepackage{amsmath}
\usepackage{amssymb}
\usepackage{multirow}
\newcommand{\tabincell}[2]{\begin{tabular}{@{}#1@{}}#2\end{tabular}}
\usepackage{subfigure}
\usepackage{wrapfig}
\usepackage{float}
\usepackage[breaklinks,colorlinks]{hyperref}
\hypersetup{citecolor=blue}
\usepackage{booktabs}
\usepackage[all]{xy}

\usepackage{newfloat}
\usepackage{listings}
\DeclareCaptionStyle{ruled}{labelfont=normalfont,labelsep=colon,strut=off} % DO NOT CHANGE THIS
\floatstyle{ruled}
\newfloat{listing}{tb}{lst}{}
\floatname{listing}{Listing}
\title{CRIN: Rotation-Invariant Point Cloud Analysis and Rotation Estimation \\ via Centrifugal Reference Frame}
\author{
    Yujing Lou\textsuperscript{\rm 1},
    Zelin Ye\textsuperscript{\rm 1},
    Yang You\textsuperscript{\rm 1}, \\
    Nianjuan Jiang\textsuperscript{\rm 2},
    Jiangbo Lu\textsuperscript{\rm 2},
    Weiming Wang\textsuperscript{\rm 1},
    Lizhuang Ma\textsuperscript{\rm 1}\thanks{Lizhuang Ma and Cewu Lu are the corresponding authors.},
    Cewu Lu\textsuperscript{\rm 1,3}\footnotemark[1]
}
\affiliations{
    \textsuperscript{\rm 1} Shanghai Jiao Tong University\ \ \ 
    \textsuperscript{\rm 2} SmartMore\ \ \
    \textsuperscript{\rm 3}Shanghai Qi Zhi Institute\\
    \{louyujing, h\_e\_r\_o, qq456cvb, wangweiming, lucewu\}@sjtu.edu.cn,  
    ma-lz@cs.sjtu.edu.cn, \\
    \{jnianjuan, jiangbo.lu\}@gmail.com
}

\usepackage{bibentry}
\begin{document}

\maketitle

\begin{abstract}
    Various recent methods attempt to implement rotation-invariant 3D deep learning by replacing the input coordinates of points with relative distances and angles. Due to the incompleteness of these low-level features, they have to undertake the expense of losing global information.
    In this paper, we propose the CRIN, namely Centrifugal Rotation-Invariant Network.
    CRIN directly takes the coordinates of points as input and transforms local points into rotation-invariant representations via centrifugal reference frames. Aided by centrifugal reference frames, each point corresponds to a discrete rotation so that the information of rotations can be implicitly stored in point features. Unfortunately, discrete points are far from describing the whole rotation space. We further introduce a continuous distribution for 3D rotations based on points. Furthermore, we propose an attention-based down-sampling strategy to sample points invariant to rotations. A relation module is adopted at last for reinforcing the long-range dependencies between sampled points and predicts the anchor point for unsupervised rotation estimation. 
    Extensive experiments show that our method achieves rotation invariance, accurately estimates the object rotation, and obtains state-of-the-art results on rotation-augmented classification and part segmentation. Ablation studies validate the effectiveness of the network design. The code is available at \href{https://github.com/yokinglou/CRIN}{https://github.com/yokinglou/CRIN}.
\end{abstract}

\input{sections/introduction.tex}

\input{sections/related_work.tex}

\input{sections/method.tex}

\input{sections/experiments.tex}

\input{sections/conclusion.tex}

\section*{Acknowledgements} This work was supported by the National Natural Science Foundation of China (72192821), National Key R\&D Program of China (2021ZD0110700), SmartMore Corporation, Shanghai Municipal Science and Technology Major Project (2021SHZDZX0102), Shanghai Qi Zhi Institute, SHEITC (2018-RGZN-02046), and Shanghai Science and Technology Commission (21511101200).

\bibliography{main}

\appendix
\section*{Appendix}
In this supplementary document, we first give the proofs of the properties reported in the main paper (Sec.~\ref{sec:proofs}). Then we further explain the rotation group equivariance used in PCRF property (Sec.~\ref{sec:equivariance}). At last, we give the object alignment procedure (Sec.~\ref{sec:obj_align}).

\input{sections/appendix.tex}

\end{document}

%% file: sections/introduction.tex
\section{Introduction}
Deep learning on point cloud has achieved tremendous development in recent years. 
Methods like PointNet\cite{qi2017pointnet}, PointNet++~\cite{qi2017pointnet++} and PointCNN~\cite{li2018pointcnn} are the pioneers for processing point cloud and obtain several achievements. However, most of these methods heavily rely on the alignment of input data. 
They always fail to generalize to unseen rotations, as the 3D object datasets~\cite{chang2015shapenet,mo2019partnet,yi2016scalable} for training always keep the objects aligned canonically. Besides, a performance decline exists after applying data augmentation since rotations are impossible to enumerate. An excess of rotation augmentation also brings unbearable computation consumption. In contrast, a rotation-invariant representation of an object is much more convenient and efficient. Therefore, seeking a novel representation for 3D objects invariant to 3D rotations is necessary.

Traditional methods~\cite{rusu2009fast,tombari2010unique} first develop hand-crafted descriptors to represent local geometries.
These descriptors utilize local geometric measurements invariant to rotations, e.g., relative distances and angles used in Point Pair Features~\cite{drost2010model,deng2018ppfnet,deng2018ppf}. However, they only focus on local geometries and ignore the global relationship.
Spherical CNNs~\cite{cohen2018spherical,esteves2018learning} first explore rotation-equivariant feature extraction. They discretize the 3D rotation group SO(3)\footnote{SO(3), the three-dimensional special orthogonal group.} on a unit sphere, extract the features for independent rotations and fuse them for global rotation-invariant features. Nevertheless, insufficient computation resources limit the resolution on the sphere, leading to an inaccurate rotation division and a considerable performance decline. 
Deep learning methods further explore the rotation invariance of point clouds. RINet~\cite{zhang2019rotation}, ClusterNet\cite{chen2019clusternet}, PR-invNet~\cite{yu2020deep} and SGMNet~\cite{xu2021sgmnet} exploit various combinations of low-level geometric measurements to replace the original coordinates of input points. Though invariant to rotations, these measurements lose essential geometric relationships from the original data. Besides, some methods~\cite{gojcic2019perfect,zhang2019rotation} build local reference frames (LRFs) by the geometric center, barycenter, etc. RI-GCN~\cite{kim2020rotation} builds LRFs by PCA of local points to transform point clouds into rotation-invariant representations. Unfortunately, these LRFs are sensitive to the distribution and density of points, which is not robust enough. Li et al.~\cite{li2021closer} achieve rotation invariance by removing the ambiguity of PCA-based canonical poses, while still requiring 24 known rotations as the prior.

To address the issues mentioned before, we propose Centrifugal Rotation-Invariant Network.
Concretely, we first introduce the Centrifugal Reference Frame (CRF), a rotation-invariant representation of point clouds. 
A CRF is based on two polar CRFs (PCRFs). The PCRF has two essential properties. First, it turns a rotation in SO(3) to a basic rotation about the $z$-axis. Second, a PCRF transforms points into a representation invariant to basic rotations about the $z$-axis of the original space. 
With the help of PCRF, a 3D rotation invariance goal is firstly simplified to a single-axis rotation invariance problem. And the remaining degree of freedom can be further eliminated by applying a PCRF again, due to its basic rotation invariance. CRF relies less on local geometry than LRFs mentioned before.
A CRF constructs one possible rotation-invariant representation. However, finding a global representation is challenging, as selecting one specific CRF from input points is difficult. To avoid this issue, instead of finding a global representation, we strive for local rotation invariance for local points in different CRFs. 

Actually, each point corresponds to a discrete rotation, as its CRF is an orthogonal matrix in SO(3). We endeavor to cover all possible rotations so that we build a continuous distribution of rotations based on input points. 
Compared with the discretization of SO(3) by predefined resolutions~\cite{cohen2018spherical}, sampling rotation from a continuous distribution is more efficient and accurate.  
CRIN also introduces an attention-based down-sampling strategy, which is robust to rotations. After down-sampling, a relation module is adopted to reinforce the long-range dependencies of points in feature space and predicts an anchor point for rotation estimation. We can estimate the rotation in an unsupervised manner via the CRF of the anchor point.
Extensive experiments show that CRIN ensures both global and local rotation invariance and achieves state-of-the-art results on rotation-augmented classification and part segmentation tasks. Besides, CRIN can estimate the rotation of an object without rotation supervision.

Finally, the main contributions of this paper are summarized as follows:
\begin{itemize}
   \item[$\bullet$] We propose a Centrifugal Reference Frame to represent the point cloud in a rotation-invariant way. 
   \item[$\bullet$] We build a continuous distribution for 3D rotations based on points and introduce a rotation-invariant down-sampling strategy. CRIN can predict an anchor point for unsupervised rotation estimation. 
   \item[$\bullet$] CRIN achieves state-of-the-art results on rotated object classification and part segmentation tasks and estimates rotations without supervision.   
\end{itemize}

%% file: sections/related_work.tex
\section{Related Work}

\paragraph{Rotation Equivariance Methods}
With the development of group convolutions~\cite{cohen2016group}, Spherical CNNs~\cite{cohen2018spherical,esteves2018learning} first propose rotation-equivariant feature extraction. Spherical CNNs discretize the 3D rotation group SO(3) on a sphere and propose a spherical convolution to extract features for each discrete rotation. These features are rotation-equivariant, and spherical CNNs fuse them to get the global rotation-invariant features. However, the resolution of the rotation group is too coarse to cover all rotations, which becomes the main obstacle for these methods.
Based on Spherical CNNs, PRIN~\cite{you2020pointwise} and SPRIN~\cite{you2021prin} make an extension to get point-wise rotation-invariant features. Equivariant Point Network~\cite{chen2021equivariant} design a group attentive pooling to fuse equivariant features into invariant counterparts, which improves the performance. Nevertheless, these methods are still constrained by the resolution of discretization. 

\paragraph{Rotation Invariance Methods}
Traditional methods design local rotation-invariant descriptors by exploiting low-level geometries. 
Hand-crafted~\cite{rusu2009fast,tombari2010unique} and learning-based descriptors\cite{deng2018ppfnet,gojcic2019perfect,khoury2017learning} both integrate local geometries into rotation-invariant features, as the local structures are invariant to the global rigid transformations. However, these descriptors lack long-range dependencies between different parts of an object. PointNet~\cite{qi2017pointnet}, SpiderCNN~\cite{xu2018spidercnn}, etc. implement a spatial transformer network (STN) to implicitly learn a transformation that aligns the input point cloud to a canonical pose. RSCNN~\cite{liu2019relation}, PointConv~\cite{wu2019pointconv}, KPConv~\cite{thomas2019kpconv}, etc.,  improve the rotation robustness with the help of STN. But STN has a drawback in that it needs numerous data augmentation to enhance the performance on rotated data. STN with data augmentation is not rigorous rotation-invariant. Besides, RIConv~\cite{zhang2019rotation}, ClusterNet~\cite{chen2019clusternet}, and SGMNet~\cite{xu2021sgmnet} replace input coordinates with low-level measurements (e.g., distances, relative angles), since the coordinate is sensitive to rotations. However, these measurements lose essential geometric information, which is deficient in recovering the original data structures. PR-invNet~\cite{yu2020deep} constructs a pose space with 120 known rotations to remove the PCA ambiguity and introduces a pose selector to find the canonical pose of an object. 
RI-GCN~\cite{kim2020rotation} proposes a local reference frame built by the stochastic dilated k-NN algorithm. Such an LRF is easily influenced by the sampling strategies and distribution of points. 
Another branch of methods~\cite{fang2020rotpredictor,spezialetti2020learning,li2021closer} tries to recover the canonical pose of the input object before feature extraction. They all need rotation augmentation or supervision during training time.

%% file: sections/method.tex
\begin{figure*}[t]
    \begin{center}
    \includegraphics[width=\linewidth]{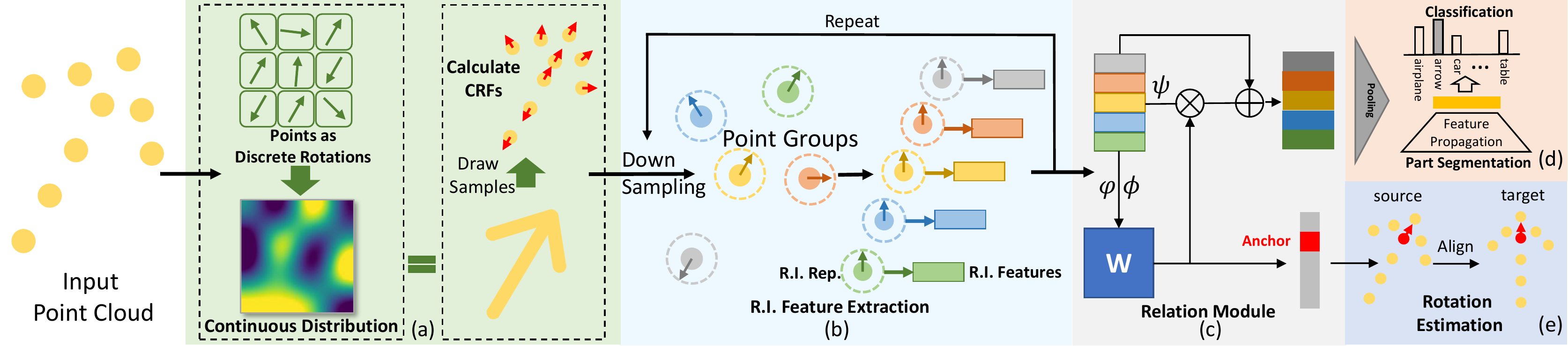}
    \end{center}
    \vspace{-10pt}
    \caption{\textbf{A 2D illustration of CRIN pipeline.} (a) CRIN builds a continuous distribution for rotations and calculates the CRFs of sampled points. (b) Down-sample points, group local points, and transform them into CRFs. Then extract local rotation-invariant features. (c) Relation module for reinforcing dependencies between points in feature space and selecting the anchor point. (d) Max-pool the global features for classification and part segmentation. (e) Rotation estimation via the anchor point.}
    \vspace{-15pt}
    \label{fig:pipeline}
\end{figure*}

\section{Method}
In this section, we first propose the Centrifugal Reference Frame (Sec.~\ref{sec:rotation_invariance}). 
Then a continuous distribution is built for 3D rotations (Sec.~\ref{sec:continuous_distribution}). 
We further propose an attention-based down-sampling method invariant to rotations and predict an anchor point to estimate the rotation of an object (Sec.~\ref{sec:sample}). 
Finally, we introduce the architecture of CRIN (Sec.~\ref{sec:architecture}).
A 2D illustration of the pipeline is demonstrated in Fig.~\ref{fig:pipeline}.

\input{sections/method/1.tex}

\input{sections/method/2.tex}

\input{sections/method/3.tex}

\input{sections/method/4.tex}

%% file: sections/method/1.tex
%-------------------------------------------------------------------------------
\subsection{Centrifugal Reference Frame}
\label{sec:rotation_invariance}
Our goal is to design a rotation-invariant representation $r:\mathbb{R}^3 \rightarrow \mathbb{R}^3$ for a point cloud $\mathcal{P}=\{p_i\}_{i=1}^N$, which satisfies $r(p)=r(Rp)$ for any $p\in\mathcal{P}$ and $R\in SO(3)$. 
Therefore, we propose the \emph{Centrifugal Reference Frame (CRF)} where points are invariant to 3D rotations. A CRF is based on an orthogonal basis $B\in\mathbb{R}^{3\times3}$, so that the representation is formulated as $r(p) = B^Tp$. 
A CRF is composed of two polar CRFs (PCRFs). We first define the PCRF.

\subsubsection{Polar Centrifugal Reference Frame} 
Given a query point $q\in\mathcal{P}$,
a PCRF is based on an orthogonal basis, denoted as $B_p=[u, v, w]\in \mathbb{R}^{3\times3}$. Specifically,
$w = \frac{q}{\|q\|}$ is defined as the centrifugal vector, 
$u = \frac{z \times w}{\|z \times w\|}$ and 
$v = w \times u$,
where $[x,y,z]\in\mathbb{R}^{3\times 3}$ is the basis of the original space. 

\begin{figure}[t]
    \begin{center}
    \includegraphics[width=\linewidth]{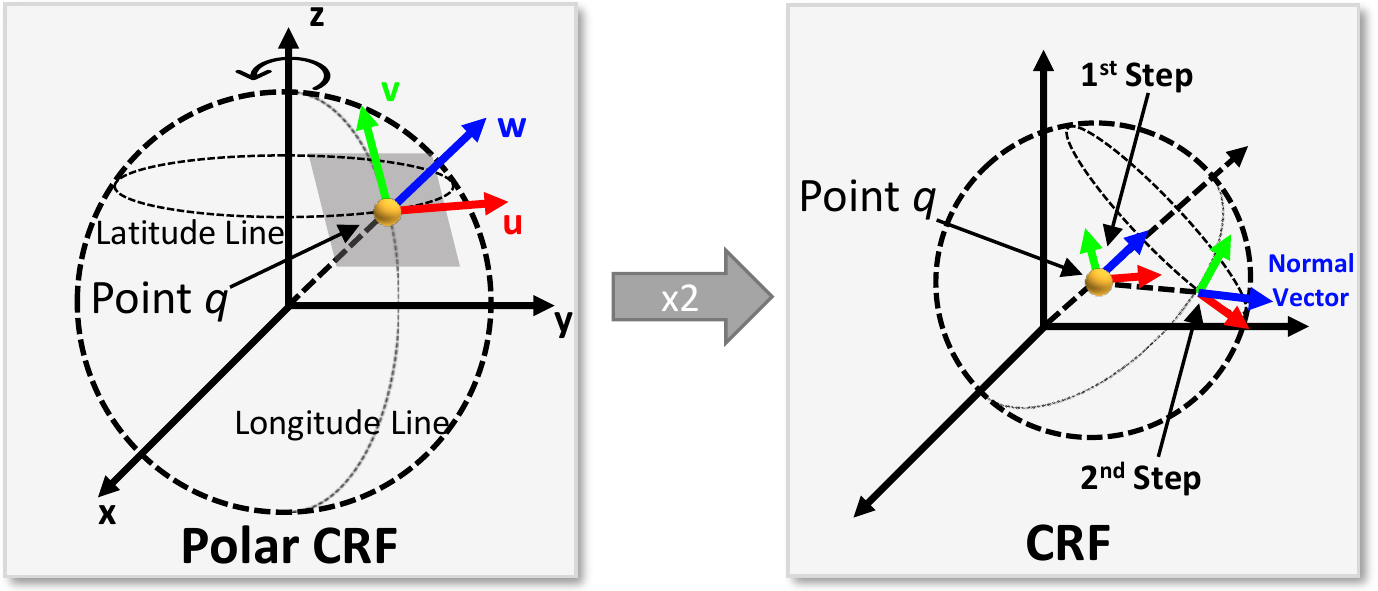}
    \end{center}
    \vspace{-10pt}
    \caption{Illustrations of CRFs.}
    \vspace{-19pt}
    \label{fig:crf}
\end{figure}

\begin{figure*}[t]
    \begin{center}
    \includegraphics[width=\linewidth]{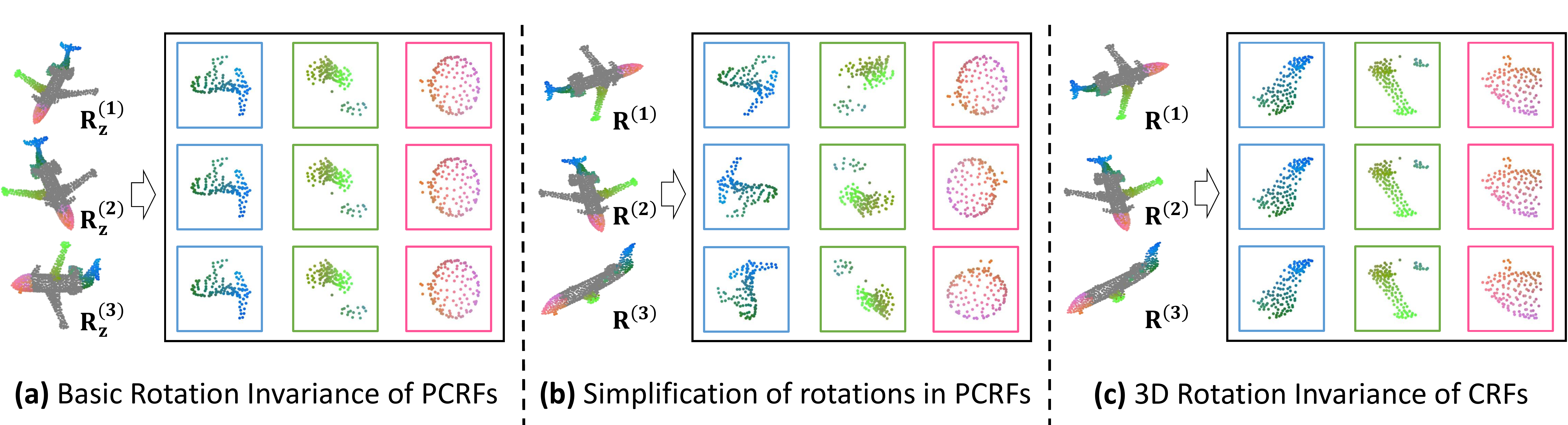}
    \end{center}
    \vspace{-13pt}
    \caption{\textbf{Illustrations of CRF properties.} Projection images of three local point groups (red, green, and blue) along the $w$-axis in different CRFs,  are given in boxes. (a) The basic rotation invariance of PCRFs. Points in PCRFs are invariant to basic rotations in the original space. (b) Simplification from 3D rotations to basic rotations.  Points revolve around the $w$-axis in PCRFs when the airplane is randomly rotated. (c) The 3D rotation invariance of CRFs.}
    \vspace{-15pt}
    \label{fig:ei}
\end{figure*}

The left image in Fig.~\ref{fig:crf} illustrates a PCRF. The name of PCRF is inspired by the characteristic that its $v$-axis is tangent to the longitude of $q$ and always points to the ``North Pole'' of the sphere. Actually, a PCRF builds a representation 
$
    g(p) = B_p^Tp,
$ 
with two essential properties:
1) PCRF builds a representation invariant to the basic rotation about the $z$-axis; 
2) PCRF simplifies a SO(3) rotation in original space to a basic rotation about the $z$-axis. 
Formally, we have
\begin{equation}
\begin{split}
    g(R_zp) = g(p),\ \ g(Rp) = R_zg(p),
\end{split}
\label{eq:pcrf}
\end{equation} 
\begin{equation}
    \begin{split}
        R_z(\theta) =
        \begin{bmatrix} 
            \cos(\theta) & -\sin(\theta) & 0 \\ 
            \sin(\theta) & \cos(\theta) & 0 \\ 
            0 & 0 & 1 \\
        \end{bmatrix}, \\
    \end{split}
    \end{equation} 
where $R_z$ is the basic rotation about the $z$-axis with a rotation angle $\theta$.  The proof of Eq.~\ref{eq:pcrf} is in the appendix~\ref{app:basic} and ~\ref{app:second_property}. The left two columns of Fig.~\ref{fig:ei} demonstrate the properties. Fig.~\ref{fig:ei}(a) shows that points in the PCRF keep static when the airplane is rotated about the $z$-axis. In Fig.~\ref{fig:ei}(b), the points revolve around the centrifugal vector in their PCRFs when the airplane is rotated by arbitrary rotations in SO(3). 
The inspiration for PCRF is from Euler's rotation theorem~\cite{doi:10.1080/07468342.1989.11973235,palais2007euler,taylor2014euler} that every 3D rotation can be specified by one axis and an angle. Cohen et al.~\cite{cohen2013learning,cohen2014learning} give that an orthogonal matrix with determinant +1 can be factorized as $R=WR_zW^T$, where $W$ is an orthogonal matrix, and $R_z$ is a basic rotation matrix about the $z$-axis. The PCRF is one specific solution of factorization.

Benefiting from the properties of PCRF, the procedure of building the rotation invariance representation can be split into two steps by applying PCRF twice. We first restrict the random rotation to one degree of freedom, i.e., the basic rotation about the centrifugal vector. Then, we eliminate the variance about the remaining axis by another PCRF. So we introduce the definition of the CRF.

\subsubsection{Centrifugal Reference Frame} 
Given a query point $q\in \mathcal{P}$, the basis of its CRF is formulated as $B = B^1_p B^2_p$, where $B^1_p$ is determined by $q$ and $B^2_p$ by taking the normal vector of $q$ as the centrifugal vector of the second PCRF, shown in the right of Fig.~\ref{fig:crf}. The CRF builds a rotation-invariant representation $r(p)$, formulated as
\begin{equation}
    r(Rp)=r(p),
\end{equation}
where $r(p) = [g^2\circ g^1](p) = B^T p$, $g^1(p)={B^1_p}^Tp$ and $g^2(p)={B^2_p}^Tp$.
The proof is in the appendix~\ref{app:ri}. Fig.~\ref{fig:ei}(c) shows that the points in the CRF are invariant to rotations in the original space.

% ClusterNet~\cite{chen2019clusternet}, PR-invNet~\cite{yu2020deep}, etc. 
Some methods~\cite{chen2019clusternet,yu2020deep} replace the input of the neural network with low-level measurements (distances, angles, etc.) to make the output rotation-invariant. However, these methods lose essential geometry information from the original data.
CRFs build the rotation-invariant representation only by changing the basis of the point cloud. The inherent geometries between points are completely reserved.  Besides, our two-step CRF relies less on local geometries and is more robust than previously mentioned LRFs, which is compared in Sec.~\ref{sec:robust}. 

Each point defines a CRF to represent the point cloud in one possible rotation-invariant way. Empirically, it is hard to select one specific CRF as the canonical representation directly. To avoid finding a specific CRF representing the whole point cloud, we turn the global rotation invariance problem into a local one. We can extract pointwise rotation-invariant features through their neighborhoods in CRFs.

%% file: sections/method/2.tex
%------------------------------------------------------------

\subsection{Continuous Distribution of Rotations}
\label{sec:continuous_distribution}

\begin{figure}[t]
    \begin{center}
    \includegraphics[width=\linewidth]{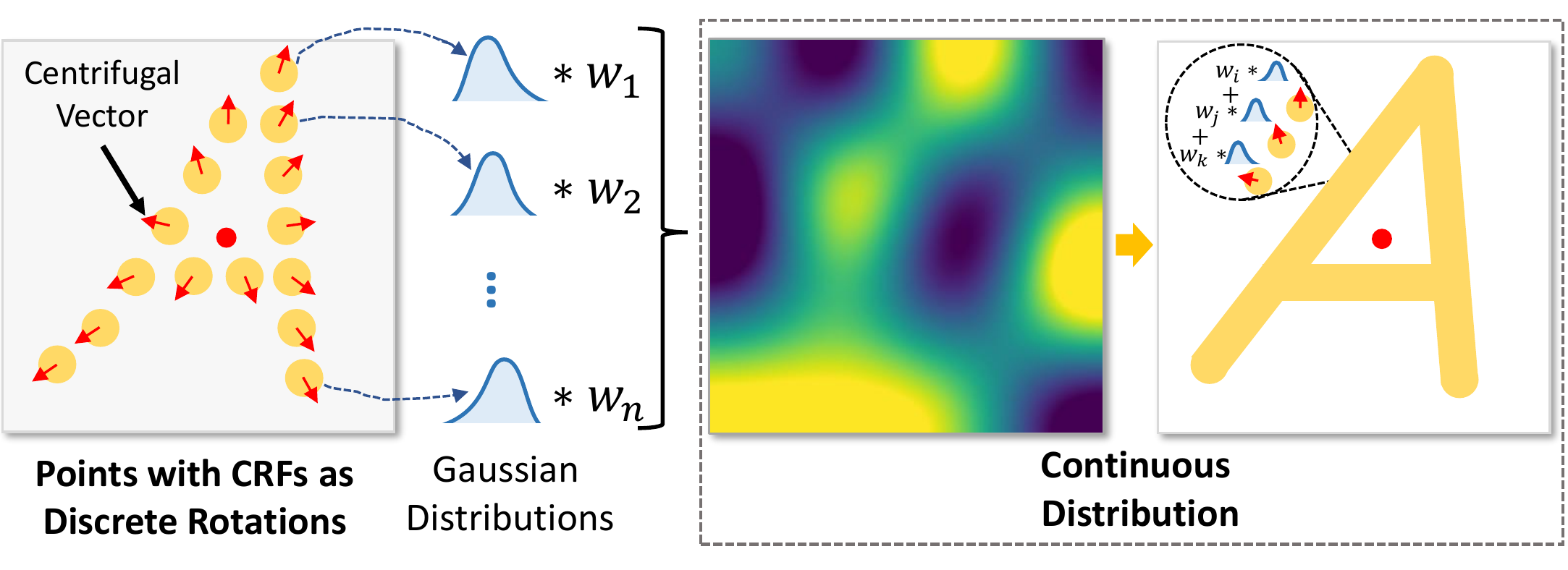}
    \end{center}
    \vspace{-15pt}
    \caption{\textbf{Procedure of building a continuous distribution in 2D.} Each point corresponds to a discrete rotation by a CRF. Gaussian distribution models each sub-distribution based on each point. The sub-distributions are summed with learnable weights to build a continuous mixture distribution.}
    \vspace{-15pt}
    \label{fig:continuous_distribution}
\end{figure}

Previous methods~\cite{chen2021equivariant,cohen2018spherical,you2020pointwise} try to discretize the rotation space with grids on a sphere and extract features for each rotation. 
Spherical CNNs~\cite{cohen2018spherical} split a sphere into grids and get rotation-equivariant features at each grid coordinate for discrete rotations. However, the discretization is inaccurate. Besides, the limited resolution leads to a vast performance decline due to its high memory occupation (see Tab.~\ref{tab:cls}). 
Actually, each CRF can be regarded as a discrete rotation since the basis of a CRF is an orthogonal matrix with the determinant $+1$. The basis is equivalent to an element in SO(3). Therefore, we map each point to a discrete rotation, as illustrated in the leftmost image of Fig.~\ref{fig:continuous_distribution}. We can further extract features for discrete points, i.e., rotations.

Considering finite discrete points are far from describing the rotation space thoroughly, we introduce a continuous distribution of rotations.
Suppose that the point cloud constructs a continuous distribution with the probability density function $f(p)$, $p\in\mathbb{R}^3$. We model it as a mixture distribution. For each input point $p_i\in\mathcal{P}$, we build a sub-distribution with the density function $f_i(p)$ using the three-variate Gaussian distribution, $\mathcal{N}(p_i, \Sigma_i)$. The covariance matrix is decided by the average closest distance between points. Therefore, the continuous distribution can be derived by a weighted sum,
$
    f(p) = \sum^{N}_{i=1} w_i f_i(p)
$,
where $w_i$ is a learnable weight.

Drawing point samples from the distribution is a two-step process. First, we sample a point $\hat{p}_i$ from $\mathcal{N}(p_i, \bm{\Sigma}_i)$ for each sub-distribution. Second, we sum all samples with learnable weights to get an element of the distribution, formulated as
$\hat{p}=\sum_{i=1}^{N}w_i\hat{p}_i$. The procedure is demonstrated in Fig.~\ref{fig:continuous_distribution}. We build the distribution and draw point samples at each entry into CRIN.
With the help of continuous distribution, we only need to sample points from the distribution instead of regressing specific rotations. 
In contrast to discretizing the rotation group, our continuous distribution covers all possible rotations. Besides, sampling rotations from distribution is more accurate and easier to implement.

%% file: sections/method/3.tex
%------------------------------------------------------------
\subsection{Unsupervised Rotation Estimation}
\label{sec:sample}
% Practically, we hope to localize a few anchor points invariant to rotation transformations to estimate the pose of an object. 
Each CRF can be used to represent one object's orientation.
We want to select an anchor point invariant to rotations from the sampled points to estimate the rotation of the point cloud. 
The hierarchical down-sampling structure is often adopted to increase the receptive fields for 3D deep learning methods. Farthest Point Sampling (FPS)~\cite{qi2017pointnet++} is always chosen because the sampled points are uniformly distributed. However, these points are not fixed because of the initial randomness, which is not qualified for rotation-invariant sampling. Although the FPS can be deterministic if we fix the initial searching point, the points sampled from the distribution vary each time, making FPS unstable. Therefore, we adopt an attention-based sampling strategy using FPS as the supervision to improve stability. Some keypoint datasets~\cite{you2020keypointnet,lou2020human} can be used to train a keypoint detector at the category level, which is unsuitable because their categories are limited.

\begin{figure}[t]
    \begin{center}
    \includegraphics[width=\linewidth]{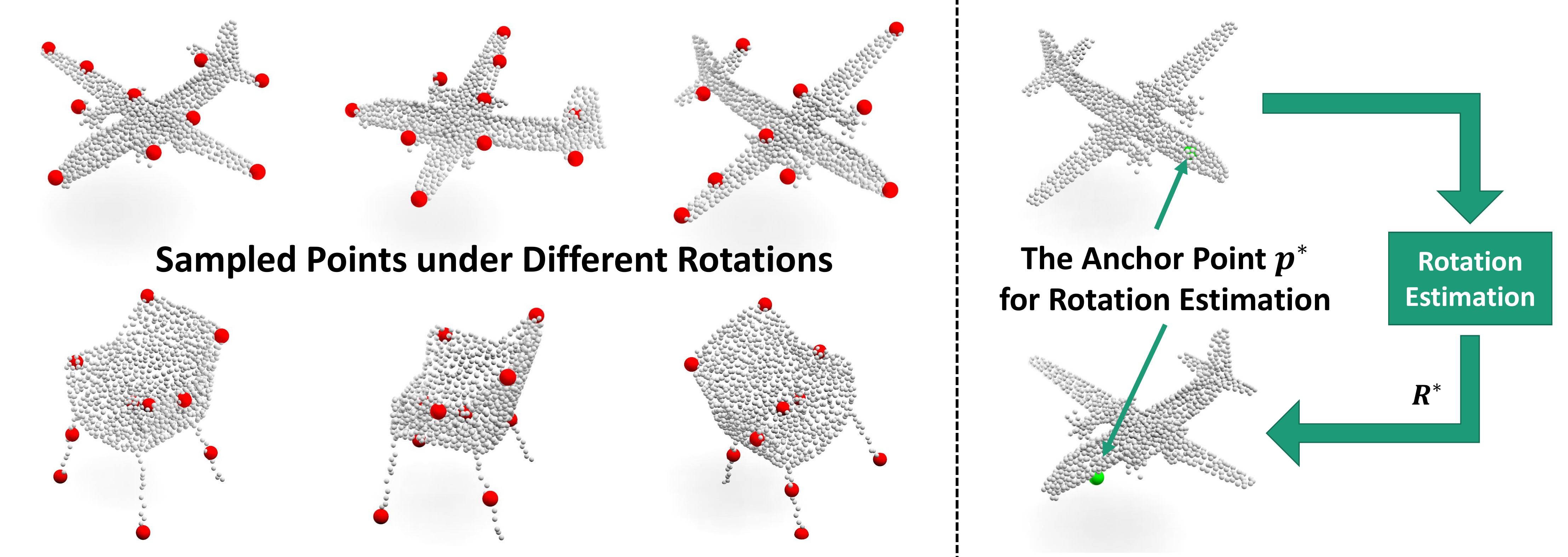}
    \end{center}
    \vspace{-15pt}
    \caption{\textbf{Left}: Points sampled by our method are robust to rotation transformations. \textbf{Right}: An anchor point is selected at the last layer for rotation estimation.}
    \vspace{-15pt}
\label{fig:sampling}
\end{figure}

Suppose the point set in layer $l$ is $\mathcal{P}_{l}$. The sampled point set $\mathcal{P}_{l+1}$ is calculated by:
\begin{equation}
    \mathcal{P}_{l+1} = softmax(\mathcal{M}(\mathcal{D})) \cdot \mathcal{P}_{l},
\end{equation}
where $\mathcal{D}$ is an $N_{l}\times N_{l}$ matrix representing distances between points in $\mathcal{P}_l$ and $N_{l}$ is point number. $\mathcal{M}$ is a multi-layer perceptron (MLP) mapping $\mathcal{D}$ to $N_{l+1}$ dimensions.
The Chamfer Distance~\cite{fan2017point} is selected for loss calculation by measuring the similarity between $\mathcal{P}_{l+1}$ and target points $\hat{\mathcal{P}}_{l+1}$ from FPS. Fig.~\ref{fig:sampling} (left) shows that points sampled in a layer by our method are stable after rotations.

After sampling points and increasing their receptive field layer-by-layer, we need to choose the anchor point at the last layer. 
Inspired by~\cite{wang2018non}, we first develop a \emph{relation module} to reinforce the dependencies between sampled points in the feature space, formulated as:
\begin{equation}
\begin{split}
    &\hat{F} = W\cdot\psi(F) + F, \\ 
    W =& softmax(\varphi(F)\cdot\phi(F)^T),
\end{split}
\end{equation}
where $F\in\mathbb{R}^{N_l\times C_l}$ is the feature matrix in layer $l$, $C_l$ is the channel number and $\varphi$, $\phi$, $\psi$ are three MLPs.
$W$ is a weight matrix to describe the dependency between points. We average matrix $W$ along the first dimension and get a feature map depicting the points' representativeness. The point $p^*$ with the highest value in the feature map is selected. The basis of the anchor CRF $B^*$ can be used to calculate the rotation. Given a point cloud with a target pose and a rotated one as the source, the estimated rotation is:
\begin{equation}
    \mathcal{R}^* = B^*_t {B^*_s}^T,
\end{equation}
where $B^*_t$ and $B^*_s$ are bases of target and source objects. 
The key to our rotation estimation is finding the anchor point, where rotation supervision is not required. 
As shown in Fig.~\ref{fig:sampling} (right), we can align the source point cloud with $\mathcal{R}^*$ to the target pose since $p^*$ is invariant to rotations.

%% file: sections/method/4.tex
\subsection{Network Architecture}
\label{sec:architecture}
Fig.~\ref{fig:pipeline} shows a 2D illustration of the CRIN architecture. It is mainly composed of five parts.
(a) CRIN first builds the continuous distribution, samples points from the distribution, and calculates the CRFs. 
(b) Then we down-sample points, group local points centered at the sampled points by k-NN algorithm~\cite{fix1989discriminatory}, and transform each group of points into corresponding CRF to get local rotation-invariant representations. 
Each group of points with features uses modified EdgeConv~\cite{dgcnn} with relative angles  to extract local rotation-invariant features.
Part (b) is repeated twice for down-sampling and increasing the receptive fields.
(c) The relation module is applied at the last layer to reinforce the relationship between points and choose the anchor point.
(d) The global rotation-invariant features are fused by max-pooling. The global features are further used for classification and part segmentation with a feature propagation module used in PointNet++~\cite{qi2017pointnet++}. 
(e) CRIN utilizes the anchor point to estimate the rotation. 
% More implementation details are in the appendix.

%% file: sections/experiments.tex
\section{Experiments}
In this section, we evaluate CRIN on several 3D object datasets and conduct the ablation study. The robustness of our CRF is also validated.
The experiments are conducted on a single GeForce RTX 2080Ti GPU and an Intel(R) Core(TM) i9-7900X @ 3.30GHz CPU.

\subsection{Object Classification}
We evaluate CRIN on ModelNet40 dataset~\cite{wu20153d} for object classification. We follow \cite{qi2017pointnet} to split the dataset into 9843 and 2468 point clouds for training and testing, respectively. Each point cloud includes 1024 points uniformly sampled from the object face and is rescaled to fit into a unit sphere. We use \textit{accuracy (\%)} over instances as the evaluation metric. We use Adam~\cite{kingma2014adam} optimizer during training and set the initial learning rate as 0.001. The batch size is 32, with about 2 minutes per training epoch on one GPU. CRIN has 1.72M parameters.

The evaluation is conducted in three different train/test settings: 1) training and testing with basic rotations about the $z$-axis (z/z); 2) training with basic rotations and testing with arbitrary rotations (z/SO(3)); 3) training and testing with arbitrary rotations (SO(3)/SO(3)). 
\vspace{-10pt}
\begin{table}[h]
    \begin{center}
        \resizebox{\linewidth}{!}{
        \begin{tabular}{l|c|c|c}
            \hline
            \textbf{Method} & \textbf{z/z} & \textbf{z/SO(3)} & \textbf{SO(3)/SO(3)}  \\
            \hline
            \hline
            PointNet~\cite{qi2017pointnet} & 89.2 & 16.4 & 75.5 \\
            PointNet++~\cite{qi2017pointnet++} & 91.8 & 18.4 & 77.4 \\
            SO-Net~\cite{li2018so} & \textbf{92.6} & 21.1 & 80.2 \\
            DGCNN~\cite{dgcnn} & 92.2 & 20.6 & 81.1 \\
            PointCNN~\cite{li2018pointcnn} & 91.3 & 41.2 & 84.5 \\
            \hline
            Spherical CNNs~\cite{cohen2018spherical} & 88.9 & 76.9 & 86.9 \\
            PRIN~\cite{you2020pointwise} & 80.1 & 70.4  & - \\
            \hline
            RIConv~\cite{zhang2019rotation} & 86.5 & 86.4 & 86.4 \\
            ClusterNet~\cite{chen2019clusternet} & 87.1 & 87.1 & 87.1 \\
            EPN~\cite{chen2021equivariant} & 88.3 & 88.1 & 88.3 \\
            PR-invNet~\cite{yu2020deep} & 89.2 & 89.2 & 89.2 \\ 
            RI-GCN~\cite{kim2020rotation} & 91.0 & 91.0 & 91.0 \\
            AECNN~\cite{zhang2020learning} & 91.0 & 91.0 & 91.0 \\
            SGMNet~\cite{xu2021sgmnet} & 90.0 & 90.0 & 90.0 \\
            LGR-Net~\cite{zhao2022rotation} & 90.9 & 90.9 & 91.1 \\
            Li et al. (w/o TTA)~\cite{li2021closer} & 90.2 & 90.2 & 90.2 \\
            Li et al. (w/ TTA)~\cite{li2021closer} & 91.6 & 91.6 & 91.6 \\
            \hline
            CRIN (ours) & 91.8 & \textbf{91.8} & \textbf{91.8} \\
            \hline
        \end{tabular}}
        \vspace{-5pt}
        \caption{Classification results on ModelNet40.}
        \vspace{-15pt}
        \label{tab:cls}
    \end{center}
\end{table}

The results are reported in Tab.~\ref{tab:cls}. It shows that our CRIN ensures rotation invariance and outperforms other networks testing in SO(3). Despite networks like SO-Net performing better in the z/z setting, they hardly approach rotation invariance. Besides, CRIN also has a better performance than other rotation-invariant methods. The results of Li et al. (w/ TTA) are close because they provide 24 rotations for augmenting the poses of PCA preprocessed objects during training and testing. Thanks to our CRF and continuous distribution, CRIN reserves the original geometries and considers all rotations. Note that there is a small branch of equivariant methods~\cite{cohen2018spherical} using NR/AR (train: no rotation / test: arbitrary rotation) setting for evaluation. CRIN also gets the same results under NR/AR setting as in Tab.~\ref{tab:cls}, as CRIN is independent of the rotation augmentation.

\subsection{Object Part Segmentation}
\begin{figure}[h]
    \centering
\includegraphics[width=\linewidth]{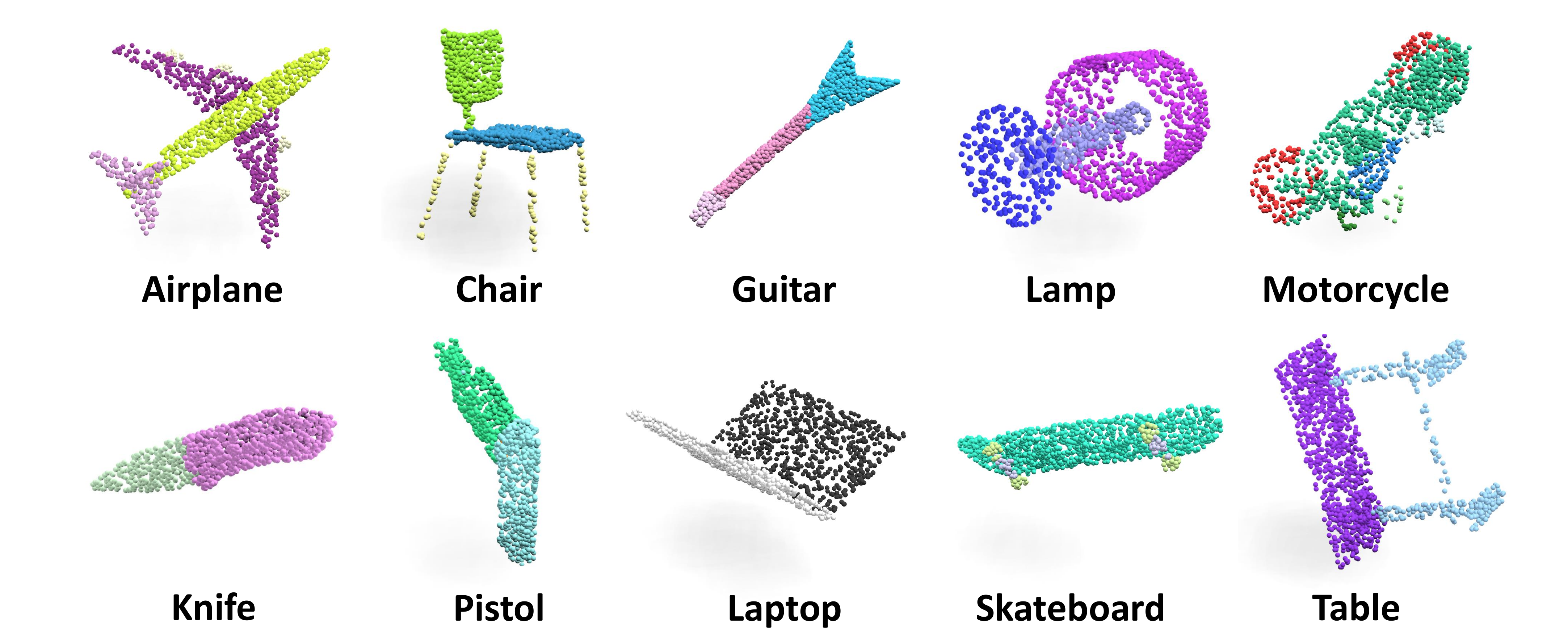}
\vspace{-20pt}
\caption{Visualizations of part segmentation.}
\vspace{-15pt}
\label{fig:ps}
\end{figure}

% \label{sec:ps}
We use the ShapeNet part dataset~\cite{yi2016scalable} for 3D part segmentation, where 16681 point clouds from 16 categories are provided. We uniformly sample 2048 points from each object. The train/test splitting is according to \cite{qi2017pointnet}. The first part of the architecture for part segmentation, which is used  to extract global features, is same as the classification task. 
We follow PointNet++~\cite{qi2017pointnet++} to propagate the global feature to input points via inverse distance weighted interpolation for per-point prediction. The evaluation metric is \textit{mean IoU scores across classes (\%)}~\cite{qi2017pointnet++}. The results are reported in Tab.~\ref{tab:ps} and validate that CRIN ensures local rotation invariance. The performances in part segmentation also support the statements in classification. Fig.~\ref{fig:ps} visualizes the part segmentation results on rotated objects.
\vspace{-10pt}
\begin{table}[t]
    \begin{center}
        \resizebox{\linewidth}{!}{\begin{tabular}{l|c|c|c}
            \hline
            \textbf{Method} & \textbf{z/z} & \textbf{z/SO(3)} & \textbf{SO(3)/SO(3)}  \\
            \hline
            \hline
            PointNet~\cite{qi2017pointnet} & 76.2 & 37.8 & 74.4 \\
            PointNet++~\cite{qi2017pointnet++} & 80.7 & 48.2 & 76.7 \\
            PointCNN~\cite{li2018pointcnn} & \textbf{81.5} & 34.7 & 71.4 \\
            DGCNN~\cite{dgcnn} & 78.8 & 37.4 & 73.3 \\
            \hline
            PRIN~\cite{you2020pointwise} & 70.3 & 54.2  & - \\
            \hline
            RIConv~\cite{zhang2019rotation} & 75.6 & 75.3 & 75.5 \\
            PR-invNet~\cite{yu2020deep} & 79.4 & 79.4 & 79.4 \\ 
            RI-GCN~\cite{kim2020rotation} & - & 77.2 & 77.3 \\
            AECNN~\cite{zhang2020learning} & 80.2 & 80.2 & 80.2 \\
            SGMNet~\cite{xu2021sgmnet} & 79.3 & 79.3 & 79.3 \\
            LGR-Net~\cite{zhao2022rotation} & 80.0 & 80.0 & 80.1 \\
            Li et al.~\cite{li2021closer} & 75.9 & 75.9 & 75.9 \\
            \hline
            CRIN (ours) & 80.5 & \textbf{80.5} & \textbf{80.5} \\
            \hline
        \end{tabular}}
        \vspace{-5pt}
        \caption{Part segmentation results on ShapeNet part dataset.}
        \vspace{-15pt}
        \label{tab:ps}
        \end{center}
\end{table}

\subsection{Classification on Real-World Object}
\begin{table}[t]
    \begin{center}
        \resizebox{\linewidth}{!}{
        \begin{tabular}{l|c|c|c}
            \hline
            \textbf{Method} & \textbf{z*/z*} & \textbf{z*/SO(3)} & \textbf{SO(3)/SO(3)}  \\
            \hline
            \hline
            PointNet~\cite{qi2017pointnet} & 79.4 & 16.7 & 54.7 \\
            PointNet++~\cite{qi2017pointnet++} & 87.8 & 15.0 & 47.4 \\
            PointCNN~\cite{li2018pointcnn} & \textbf{89.9} & 14.6 & 63.7 \\
            DGCNN~\cite{dgcnn} & 87.3 & 17.7 & 71.8 \\
            \hline
            RIConv~\cite{zhang2019rotation} & - & 78.4 & 78.1 \\
            LGR-Net~\cite{zhao2022rotation} & - & 81.2 & 81.4 \\
            Li et al. (w/o TTA)~\cite{li2021closer} & 84.3 & 84.3 & 84.3 \\
            Li et al. (w/ TTA)~\cite{li2021closer} & 86.7 & \textbf{86.7} & \textbf{86.7} \\
            \hline
            CRIN (ours) & 84.7 & 84.7 & 84.7 \\
            \hline
        \end{tabular}}
        \vspace{-5pt}
        \caption{Classification results on ScanObjectNN.}
        \vspace{-20pt}
        \label{tab:cls_scan}
    \end{center}
\end{table}
We also test CRIN on the real-world dataset ScanObjectNN~\cite{uy-scanobjectnn-iccv19}. We follow Li et al.~\cite{li2021closer}, using the OBJ\_BG split of ScanObjetNN, which includes 15 categories and 2890 indoor objects, where 2312 for training and 578 for testing. 
And z* denotes objects without any rotations. 
As shown in Tab.~\ref{tab:cls_scan}, CRIN also performs well on real-world objects. The performance of Li et al. (w/ TTA) is better, benefiting from 24 rotations augmentation during both training and testing. CRIN, without any rotation augmentation, even outperforms Li et al. (w/o TTA) with augmentation during training.

\subsection{Object Rotation Estimation}

We conduct rotation estimation on the ModelNet40.  For the evaluation metric, we adopt the \textit{Average Distance (AD)} metric used in 6D pose estimation~\cite{xiang2018posecnn}. Given an aligned point cloud $\mathcal{P}$ with $N$ points, we randomly rotate the point cloud $K$ times, and the rotated point clouds are denoted as $\{\hat{\mathcal{P}}_k\}_{k=1}^K$. Suppose the predicted rotations are $\{\hat{R}_k\}_{k=1}^K$. The average distance computes the mean of the pairwise distances:
\begin{equation}
    AD = \frac{1}{KN}\sum_{k=1}^K\sum_{i=1}^N \| p_i - \hat{R}_k \cdot \hat{p}_{ki} \|_2,
\end{equation}
where $p_i\in \mathcal{P}$ and $\hat{p}_{ki} \in \hat{\mathcal{P}}_k$. The pose is considered to be correct if the average distance is smaller than 10\% of the 3D object diameter,  following~\cite{xiang2018posecnn}. 

We compare CRIN with three types of methods. The first one is the Iterative Closest Point (ICP)~\cite{rusinkiewicz2001efficient} algorithm, which minimizes the similarity between point clouds iteratively. We use the implementation in Open3D~\cite{Zhou2018} with 2000 maximum iterations. The second type is estimating the rotation by calculating the descriptors of points and using RANSAC~\cite{fischler1981random} to find the correspondences between two point clouds. The third type is learning to estimate the rotation. We train the point cloud head of DenseFusion~\cite{wang2019densefusion} and EPN~\cite{chen2021equivariant} with rotation supervision on ModelNet40.

\vspace{-5pt}
\begin{table}[h]
    \begin{center}
    \resizebox{\linewidth}{!}{
    \begin{tabular}{l|cc}
        \hline
        \textbf{Method}  & \textbf{Mean AD $\downarrow$} & \textbf{Accuracy $\uparrow$} \\
        \hline
        \hline
        ICP~\cite{rusinkiewicz2001efficient}  & 0.6965 & 0.2 \\
        \hline
        FPFH~\cite{rusu2009fast} & 0.0968 & 84.1 \\
        SHOT~\cite{tombari2010unique} & 0.0699 & 93.6 \\
        CGF~\cite{khoury2017learning} & 0.0700 & 92.0 \\
        SpinNet~\cite{ao2021spinnet} & \textbf{0.0672} & 95.2 \\
        \hline
        DenseFusion~\cite{wang2019densefusion}  & 0.0689 & 96.4 \\
        EPN~\cite{chen2021equivariant} & 0.0684 & 98.1 \\
        \hline
        CRIN(ours) & 0.0678 & \textbf{98.9} \\
        \hline
    \end{tabular}}
    \end{center}
    \vspace{-10pt}
    \caption{Rotation estimation results on ModelNet40.}
    \vspace{-10pt}
    \label{tab:pose}
\end{table}

\begin{figure}[ht]
    \centering
\includegraphics[width=\linewidth]{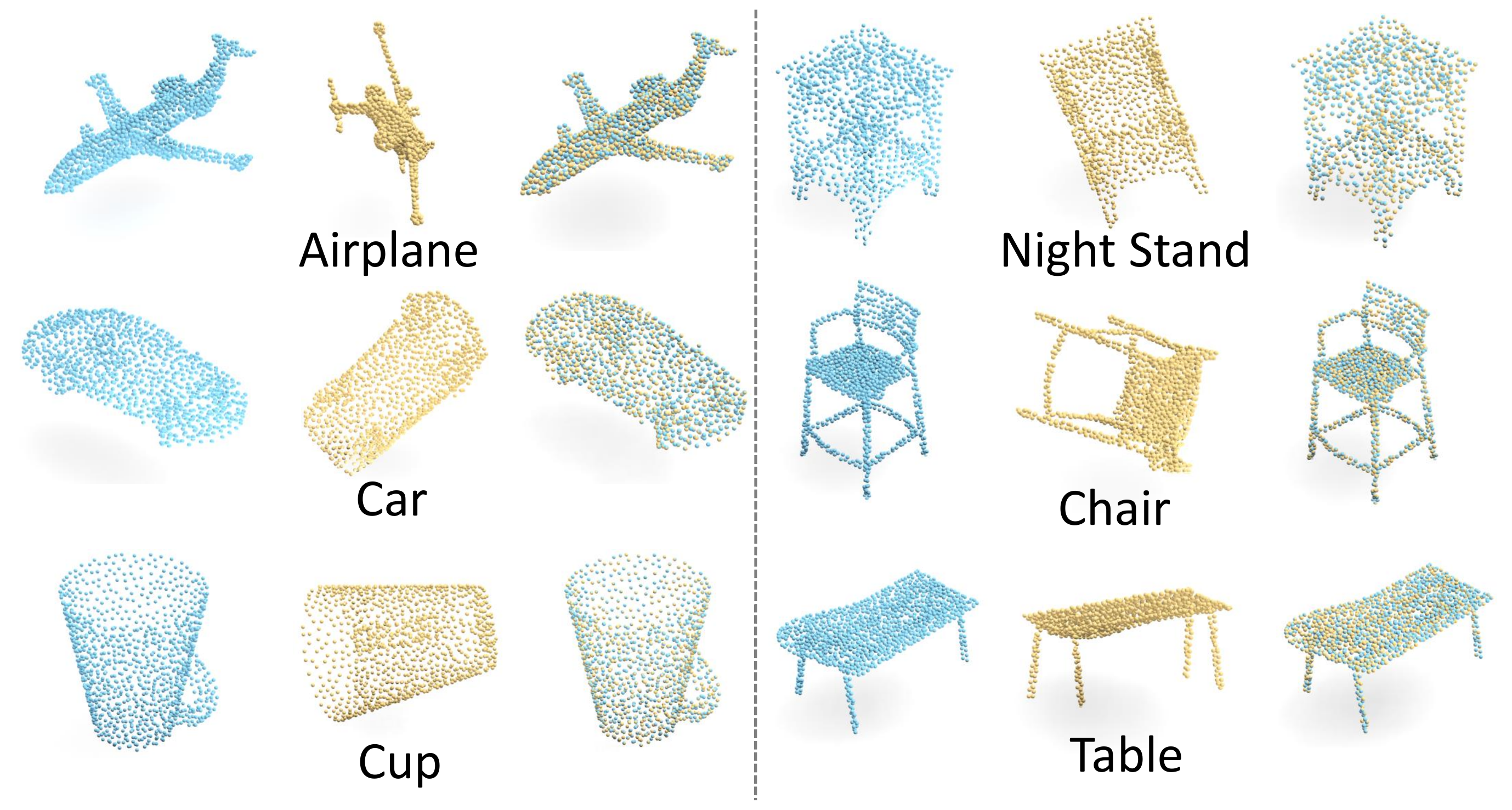}
\vspace{-20pt}
\caption{Visualizations of rotation estimation results.}
\vspace{-10pt}
\label{fig:align}
\end{figure}

The test set of ModelNet40 is used for evaluation. The rotation number $K$ is set to 16. To validate the robustness of rotation estimation, we add the Gaussian noise with the standard deviation 0.01.
The mean AD and accuracy (\%) across 40 classes are listed in Tab.~\ref{tab:pose}. 
ICP aligns objects rotated with large angles in a totally wrong orientation, as the objective of ICP is to align the object to a pose with the largest overlap with the target.  Finding correspondences by descriptors and RANSAC can improve the results, while these methods consume excessive time to match two point clouds. Compared with learning methods, CRIN outperforms them even without rotation supervision.
The visualization of alignment results is shown in Fig.~\ref{fig:align}. Objects in blue are in target poses and objects in yellow are randomly rotated. The alignment results are shown in the third column.
Objects aligned by our CRIN almost coincide with the target objects. 
Therefore, we have validated that the anchor point predicted by CRIN can estimate the pose of the object.

\subsection{Ablation Study}
\label{sec:ablation}
We conduct ablation studies to validate the design of CRIN. The results are shown in Tab.~\ref{tab:ablation}.

% \vspace{-5pt}
\begin{table}[ht]
    \begin{center}
        \resizebox{0.8\linewidth}{!}{
            \begin{tabular}{cccc|cc}
                \hline
                \textbf{CRF} & \tabincell{c}{\textbf{Continuous}\\\textbf{Distribution}} & \tabincell{c}{\textbf{Relation}\\\textbf{Module}} & \tabincell{c}{\textbf{Attn.-based}\\\textbf{Sampling}} & \textbf{z/z} & \textbf{z/SO(3)}  \\
                \hline
                \hline
                - & - & - & - & 90.7 & 16.8 \\
                polar & - &  - & - & 88.8 & 77.0 \\
                $\surd$ & -  &  - & - & 89.1 & 89.1 \\
                $\surd$ & -  & - & $\surd$ & 89.3 & 89.3 \\
                $\surd$ & -  & $\surd$ & - & 90.0 & 90.0 \\
                $\surd$ & $\surd$  &  - & - & 90.9 & 90.9 \\
                $\surd$ & $\surd$  &  $\surd$ & $\surd$ & \textbf{91.8} & \textbf{91.8} \\
                \hline
            \end{tabular}
        }
    \end{center}
    \vspace{-10pt}
    \caption{Ablation study.}
    \vspace{-15pt}
    \label{tab:ablation}
\end{table}

The baseline performs well in the z/z setting. However, it is sensitive to rotations. The PCRF narrows the gap between results in z/z and z/SO(3), which is not true rotation invariance. It just constrains the 3D rotation to one degree of freedom. The augmentation of basic rotations about the $z$-axis during training compensates for the information about the remaining degree of freedom while is not good enough. Hence, a complete CRF ensures rigorous rotation invariance.
Despite improving minor, the attention-based sampling ensures rotation robust sampling.
The continuous distribution makes up for the deficiency of discrete rotations, improving the performance by a remarkable margin. 
The relation module increases the dependencies between sampled points. The improvement by the relation module is less than other modules because the receptive fields of anchor points are partially overlapped. Part of the relations has been included during the down-sampling process.

%-------------------------------------------------------------------------------

\begin{table}[t]
    \begin{center}
    \resizebox{\linewidth}{!}{
    \begin{tabular}{l|cccccccccc|cc}
        \hline
        \textbf{Method}  & \textbf{0.0} & \textbf{0.1} & \textbf{0.2} & \textbf{0.3} & \textbf{0.4} & \textbf{0.5} & \textbf{0.6} & \textbf{0.7} & \textbf{0.8} & \textbf{0.9} & \textbf{avg. $\uparrow$} & \textbf{var. $\downarrow$} \\
        \hline
        \hline
        RI-GCN & 89.2 & 86.1 & 83.2 & 62.7 & 16.0 & 6.9 & 5.1 & 4.6 & 3.6 & 3.2 & 36.0 & 1359.9 \\
        RIConv & 86.5 & 85.0 & 82.3 & 80.3 & 76.3 & 70.1 & 63.0 & 57.0 & 50.8 & 45.7 & 69.7 & 197.0 \\
        Cov. & 89.6 & 88.9 & 88.6 & 88.1 & 87.9 & 88.2 & 87.1 & 86.7 & 75.8 & 71.1 & 85.2 & 36.2 \\
        AECNN & 90.9 & 90.4 & 90.2 & 90.0 & 89.6 & 89.7 & 88.9 & 88.3 & 82.6 & 78.1 & 87.9 & 15.6 \\
        \hline
        CRIN & 91.8 & 91.4 & 91.5 & 91.4 & 91.3 & 91.0 & 90.4 & 90.6 & 88.1 & 86.3 & \textbf{90.4} & \textbf{2.9} \\
        \hline
    \end{tabular}}
    \end{center}
    \vspace{-10pt}
    \caption{Robustness analysis of different LRFs.}
    \vspace{-20pt}
    \label{tab:robust}
\end{table}

\subsection{Robustness Analysis}
\label{sec:robust}

\noindent\textbf{Robustness to Point Density}\ \ Our two-step representation, CRF, is robust to local structural changes. To validate it, we compare CRF with other rotation-invariant LRFs: 1) RI-GCN~\cite{kim2020rotation}, LRF based on PCA of local points; 2) RIConv~\cite{zhang2019rotation}, low-level features estimated by the barycenter, geometric center of local points and the origin; 3) LRF based on the covariance matrix of local points~\cite{gojcic2019perfect}; 4) AECNN~\cite{zhang2020learning}, LRF estimated by centrifugal vector and the barycenter of local points. Tab.~\ref{tab:robust} lists different method results tested with the point dropout ratio from 0 to 0.9. It shows that CRIN is more independent of local points.
Actually, these LRFs rely heavily on local structure. The local structure perturbation easily influences their performances. CRF utilizes two PCRFs to avoid over-reliance on local points. Besides, the processing times for a batch of point clouds are 1.74ms (CRF), 8.30ms (RIConv), 7.85ms (AECNN), and 8.27ms (Cov.). CRF is more efficient, as CRF is free of grouping points or calculating centers, distances, etc.

\begin{figure}[ht]
    \centering
\includegraphics[width=\linewidth]{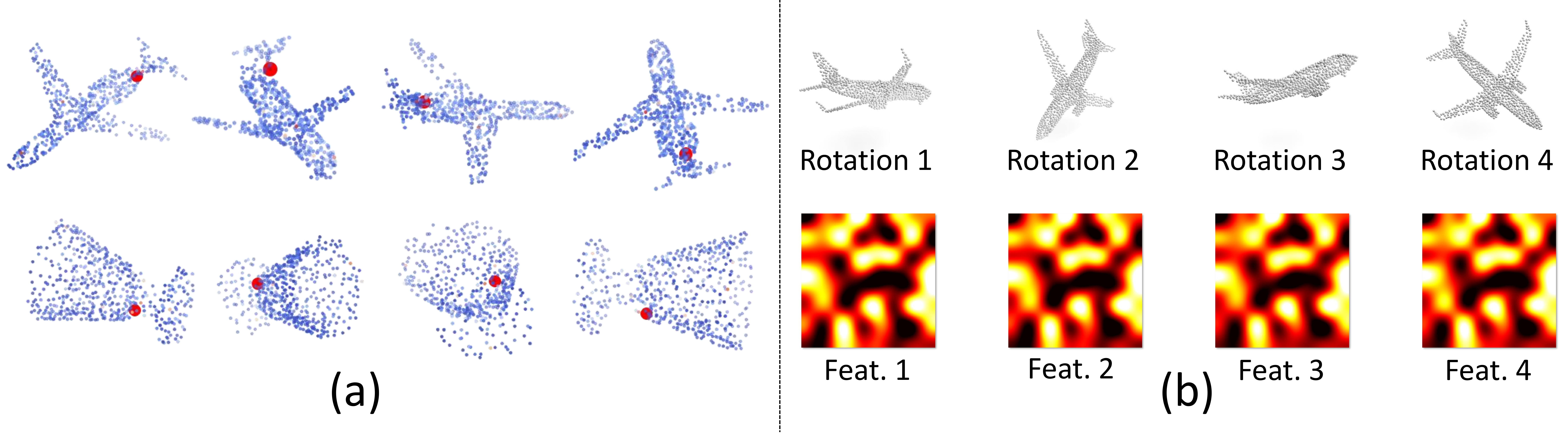}
\vspace{-20pt}
\caption{Visualization of anchor points and point features.}
\vspace{-15pt}
\label{fig:ablation}
\end{figure}

\noindent\textbf{Anchor Point for Rotation Estimation}\ \  
To ensure the robustness of anchor point selection, we visualize the anchor point used for rotation estimation under different rotations. In Fig.~\ref{fig:ablation}(a), the anchor point (red) used for rotation estimation is relatively fixed on the objects with different poses.

\noindent\textbf{Rotation-Invariant Features}\ \  To validate CRIN predicts per-point rotation-invariant features, we visualize point features in Fig.~\ref{fig:ablation}(b). We select 64 points uniformly from the point cloud and rearrange their features into an $8\times 8$ matrix. We can see that these features almost keep same during rotation transformations.

\vspace{-10pt}

%% file: sections/conclusion.tex
\section{Conclusion and Limitation}
We present the CRIN for rotation invariance analysis of 3D point cloud. We introduce the centrifugal reference frame as a rotation-invariant representation of the point cloud, which reserves the input data structure without losing geometry information.
To avoid localizing one specific CRF, we turn the global rotation invariance to a local one. Each point with CRF can be treated as a discrete rotation, and a continuous distribution of rotation space is further built based on points. 
Furthermore, CRIN utilizes a down-sampling strategy robust to rotations and a relation module to reinforce the relationship between sampled points in feature space. The relation module at the last layer predicts an anchor point for unsupervised rotation estimation. Experiments show that CRIN is qualified for rotation-invariant point cloud analysis. 

In real-world applications, our CRIN is inevitably influenced by the shift of the global center due to the occlusion or the background points. A straightforward solution to this issue is applying the whole CRIN on each local patch with the local geometric center as the origin. Therefore, we can get rotation-invariant features of the local patch, regardless of the global center. The global rotation-invariant features can be fused with all local features. In addition, some center voting strategies~\cite{you2022cppf,lin2022sar} also can improve the robustness of CRIN. We will focus on rotation invariance in real-world applications in our future work.

%% file: sections/appendix.tex
\section{Proofs of CRF Properties}
\label{sec:proofs}

\subsection{Basic Rotation Invariance in PCRF}
\label{app:basic}
\begin{figure}[h]
\begin{center}
   \includegraphics[width=0.7\linewidth]{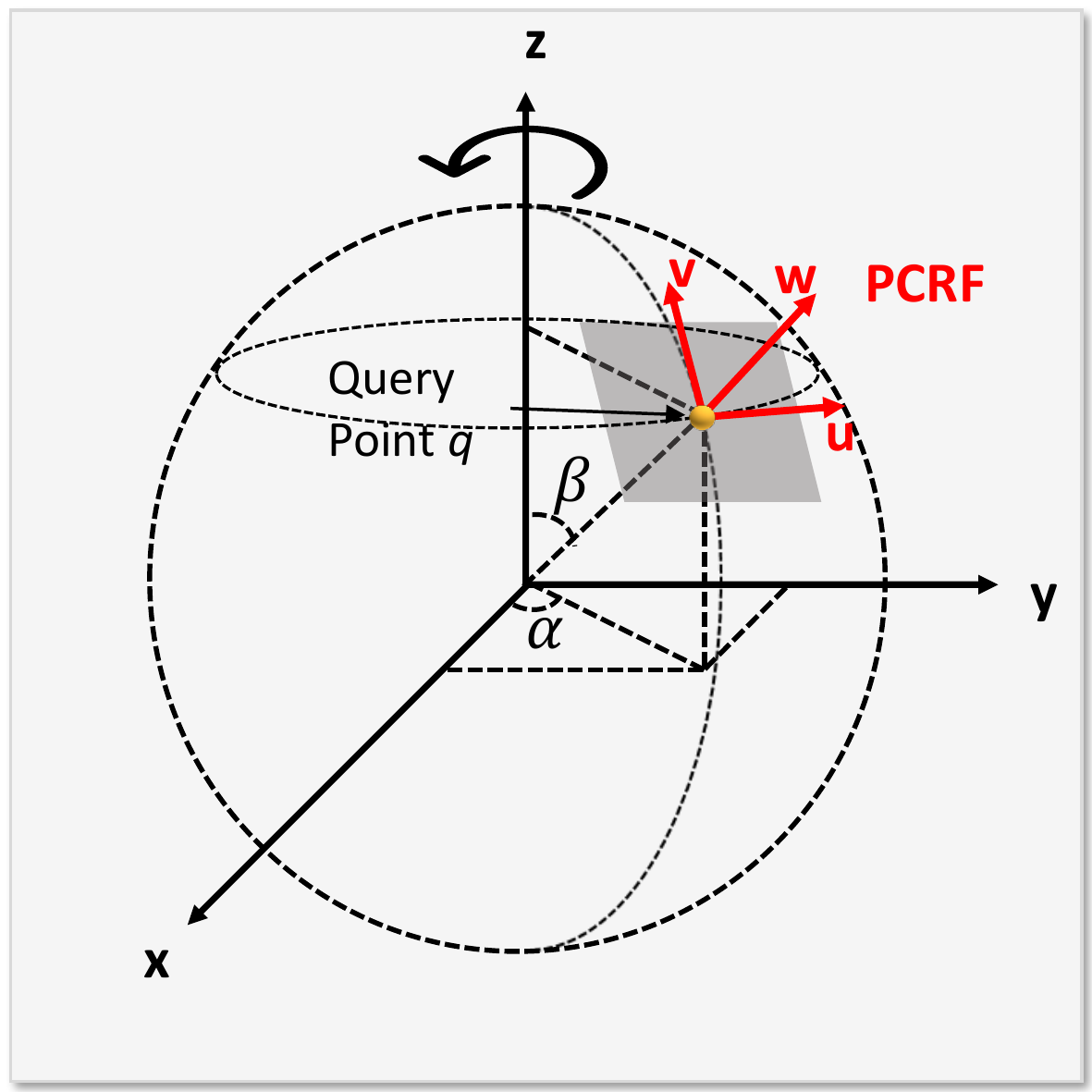}
\end{center}
   \caption{An illustration of a PCRF structure.}
\label{fig:pcrf}
\end{figure}

We first review the basic rotation invariance of a PCRF. The PCRF of a query point $q\in\mathcal{P}$ builds a basic rotation invariant representation $g(p)=B_p^Tp$, which satisfies that:
\begin{equation}
    g(R_zp) = g(p),
\end{equation}
where $p\in\mathcal{P}$.

\noindent\textbf{Proof}\ \  Suppose that the basis of PCRF is $B_p=[u,v,w]$, where the unit centrifugal vector $w=[w_1, w_2, w_3]^T$. Figure~\ref{fig:pcrf} illustrates a PCRF, where $\alpha$ and $\beta$ are the azimuthal and polar angles of $w$. The basis is subject to:
\begin{equation}
\begin{split}
    &\|w\| = 1, \\
    &u=[-\sin\alpha, \cos\alpha,0]^T, \\
    &v=[-\cos\alpha\cos\beta,-\sin\alpha\cos\beta,\sin\beta]^T, \\
\end{split}
\end{equation}
where
\begin{equation}
\begin{split}
    \sin\alpha &= \frac{w_2}{(w_1^2+w_2^2)^{\frac{1}{2}}},\quad \cos\alpha = \frac{w_1}{(w_1^2+w_2^2)^{\frac{1}{2}}}, \\
    \sin\beta &= (w_1^2+w_2^2)^{\frac{1}{2}},\quad \cos\beta = w_3. \\
\end{split}
\end{equation}
Note that these equations of the rotated $q$ have the same formulations. 
We use $w'$ to represent the rotated version of $w$, and the rest is deduced by analogy. 
Specifically, the centrifugal vector of the rotated point is $w' = R_z w$.

Suppose that the whole point cloud is rotated by a basic rotation $R_z$ about the $z$-axis with an angle $\theta$, where 
\begin{equation}
    R_z(\theta) = 
    \begin{bmatrix} 
        \cos(\theta) & -\sin(\theta) & 0 \\ 
        \sin(\theta) & \cos(\theta) & 0 \\ 
        0 & 0 & 1 \\
    \end{bmatrix}.
    \end{equation}
For any point $p=[p_1, p_2, p_3]^T\in \mathcal{P}$ and basic rotation $R_z$, we have 
\begin{equation}
\begin{split}
    g&(R_zp) = B_p'^T R_z p \\
    =& [u', v', w']^T R_z p \\
    =& \begin{bmatrix} 
    -\sin\alpha' & -\cos\alpha'\cos\beta' & w_1' \\
    \cos\alpha' & -\sin\alpha'\cos\beta' & w_2' \\
    0 & \sin\beta' & w_3' 
    \end{bmatrix}^T \\
    &\begin{bmatrix}
        \cos\theta & -\sin\theta & 0 \\ 
        \sin\theta & \cos\theta & 0 \\ 
        0 & 0 & 1 
    \end{bmatrix}
    \begin{bmatrix}
        p_1 \\
        p_2 \\
        p_3
    \end{bmatrix}\\
    =& \begin{bmatrix}
        \frac{-w_2}{(w_1^2+w_2^2)^{\frac{1}{2}}} & \frac{w_1}{(w_1^2+w_2^2)^{\frac{1}{2}}} & 0\\ -\frac{w_1w_3}{(w_1^2+w_2^2)^{\frac{1}{2}}} & -\frac{w_2w_3}{(w_1^2+w_2^2)^{\frac{1}{2}}} & (w_1^2+w_2^2)^{\frac{1}{2}}\\ 
        w_1 & w_2 & w_3 
    \end{bmatrix}
    \begin{bmatrix}
        p_1 \\
        p_2 \\
        p_3
    \end{bmatrix}\\\\
    =& \begin{bmatrix} 
    -\sin\alpha & -\cos\alpha\cos\beta & w_1 \\
    \cos\alpha & -\sin\alpha\cos\beta & w_2 \\
    0 & \sin\beta & w_3
    \end{bmatrix}^T
    \begin{bmatrix}
        p_1 \\
        p_2 \\
        p_3
    \end{bmatrix}\\\\
    =& B_p^Tp \\
    =& g(p)
\end{split}
\label{eq:proof_pcrf}
\end{equation}
$B_p'^T$ used in the first line of Eq.~\ref{eq:proof_pcrf} is the basis of rotated $q$, since the PCRF of rotated $q$ is changed after the whole point cloud $\mathcal{P}$ is rotated by $R_z$.
Here we have proved the basic rotation invariance of a PCRF.

\subsection{The Second Property of PCRF}
\label{app:second_property}
The second property is that a PCRF of query point $q\in\mathcal{P}$ constructs a representation satisfied that:
\begin{equation}
    g(Rp)=R_zg(p).
\label{eq:property_bcrf}
\end{equation}

Before proving Eq.~\ref{eq:property_bcrf}, we first introduce a theorem. Cohen et al.~\cite{cohen2013learning,cohen2014learning} introduce that an orthogonal matrix with determinant +1, i.e. the elements in the special orthogonal group, can be factorized as 
\begin{equation}
    R=WR_zW^T, 
\label{eq:factorize}
\end{equation}
where $W$ is an orthogonal matrix and $R_z$ is a basic rotation matrix about the $z$-axis.
Eq.~\ref{eq:factorize} gives that a 3D rotation matrix is equivalent to a basic rotation matrix about the $z$-axis, since matrix $W$ is invertible. The 3D rotation matrix can be factorized by a change of basis. It builds a relationship between 3D rotations and basic rotations. 

To prove the property of PCRF, we need a lemma of Eq.~\ref{eq:factorize}.
Left multiply a rotation matrix $R'\in SO(3)$ on both sides of Eq.~\ref{eq:factorize}, we have $R'R = R'WR_zW^T$, where $R'R$ is still a rotation matrix and $R'W$ is a new orthogonal matrix. Then we have 
\begin{equation}
    R = W' R_zW^T,
\label{eq:lemma}
\end{equation}
where $W'$ and $W$ are two different $3\times3$ orthogonal matrices

In our case, we use the orthogonal basis of the PCRF to build the linear mapping between an arbitrary 3D rotation and a basic rotation. In other words, an arbitrary 3D rotation can be mapped to a basic rotation by changing the basis of PCRF, formulated as:
\begin{equation}
\begin{split}
    R &= B'_p R_z B_p^T, \\
    R_z &= B'^T_p R B_p,
\end{split}
\label{eq:cor}
\end{equation}
where $B_p$ is an orthogonal basis of point $q$'s PCRF and $B'_p$ is an orthogonal basis of the rotated $q$.
With the theorem foundations introduced before, we give the proof of Eq.~\ref{eq:property_bcrf}.

\noindent\textbf{Proof}\ \ For any $p\in\mathcal{P}$ and rotation $R\in SO(3)$, we have:
\begin{equation}
\begin{split}
    g(Rp)&= B_p'^T R p  \\
    &= B_p'^T (B_p' R_z B_p^T) p \quad{\rm (Eq.~\ref{eq:cor})} \\
    &= (B_p'^T B_p') R_z B_p^T p \\
    &= R_z B_p^T p \\
    &= R_z g(p),
\end{split}
\label{eq:proof_bcrf}
\end{equation}
where $B_p$ is the basis of point $q$'s PCRF and $B_p'$ is the basis of the rotated point $Rq$.
Therefore, we have proved the property of a PCRF.
Actually, the property of PCRF include a group equivariance between $SO(3)$ and its subgroup. See more details in appendix~\ref{sec:equivariance}.

\subsection{Rotation Invariance in CRF}
\label{app:ri}
CRF builds a rotation-invariant representation of the point cloud, which satisfies the following:
\begin{equation}
    r(Rp) = r(p).
\end{equation}

\noindent\textbf{Proof}\ \  
Suppose that the basis $B$ of a CRF is combined with two PCRFs based on $B^1_p$ and $B^2_p$. 
For any point $p\in\mathcal{P}$ and rotation $R \in SO(3)$, we have:
\begin{equation}
\begin{split}
    r(Rp) &= B'^T R p \\
    &= {B^{2}_p}'^T {B^{1}_p}'^T R p \\
    &= {B^{2}_p}'^T (R_z {B^{1}_p}^Tp) \quad {\rm (Eq.~\ref{eq:proof_bcrf})} \\
    &= {B^{2}_p}^T {B^{1}_p}^T p \quad {\rm (Eq.~\ref{eq:proof_pcrf})}\\
    &= (B^1_p B^2_p)^T p\\
    &= B^Tp \\
    &= r(p).
\end{split}
\end{equation}
$B'$ is the CRF basis of rotated point $Rq$.
Here we have proved the 3D rotation invariance of a CRF.

% ------------------------------------------------------------------------------------------------------------
\section{Rotation Group Equivariance}
\label{sec:equivariance}
The concept for group equivariance is referred from Group Equivariant Convolutional Networks~\cite{cohen2016group}. A mapping $\Phi$ is equivariant s.t.
\begin{equation}
    \Phi(T_gx) = T_g'\Phi(x),\ g\in G.
\end{equation}
$T_g$ transforms $x$ with a transformation $g$ (forming $T_gx$). $T$ and $T'$ are linear representations of G and need not be same.

Actually, the basic rotation $R_z$ set with the binary operation ``$\cdot$'' of SO(3) group constructs a subgroup of SO(3). Let $(H, \cdot)$ represent the group of the basic rotations. For any $h_1,h_2\in H$, we have $h_1\cdot h_2 \in H$ and $h_1^{-1} \in H$, since the multiplication of two basic rotations about the $z$-axis is a basic rotation and the inverse of a basic rotation is a basic rotation. Therefore, $H$ is a subgroup of $SO(3)$ (i.e. $H\leq SO(3)$). 

In our setting, the equivariant mapping is $\eta: SO(3) \rightarrow H$. Concretely, $\eta(x)=W'^{-1} x W$ (according to Eq.~\ref{eq:lemma}), where $x\in SO(3)$. 
For any $R\in SO(3)$, we have 
\begin{equation}
    \begin{split}
        \eta(R x) &= W'^{-1} R x W \\
        &= W'^{-1} (W' R_z W'^T) x W \quad {\rm (Eq.~\ref{eq:factorize})} \\
        &= R_z W'^{-1} x W \\
        &= R_z \eta(x),
    \end{split}
    \label{eq:equivariance}
\end{equation}
where $R_z \in H$.  Besides, the equivariance is held when $W'=W$. In Eq.~\ref{eq:equivariance}, two transformation operators are
\begin{equation}
    \begin{split}
        T_{R}&: SO(3) \rightarrow SO(3), \\
        T'_{R}&: H \rightarrow H.
    \end{split}
\end{equation}
$T$ is obviously a linear representation of $SO(3)$. $T'$ is a linear representation because of the factorization introduced in Eq.~\ref{eq:factorize}, i.e. $R_z = W^{-1} R W$. Then we have
\begin{equation}
    \eta(T_Rx) = T'_R\eta(x).
\end{equation}
The rotation group equivariance is illustrated as:
\begin{equation}
    \xymatrix{
        SO(3)\ar[d]^\eta \ar[r]^{T_R} & SO(3)\ar[d]^\eta \\
        H\ar[r]^{T'_R} & H. \\
    }   
\end{equation}
Therefore, we have built an equivariant mapping from $SO(3)$ and $H$. Based on this equivariant mapping, we can further draw the property of our PCRF (Eq.~\ref{eq:property_bcrf}).

% ------------------------------------------------------------------------------------------------------------
\section{Object Alignment}
\label{sec:obj_align}
\begin{figure}[t]
    \begin{center}
       \includegraphics[width=0.8\linewidth]{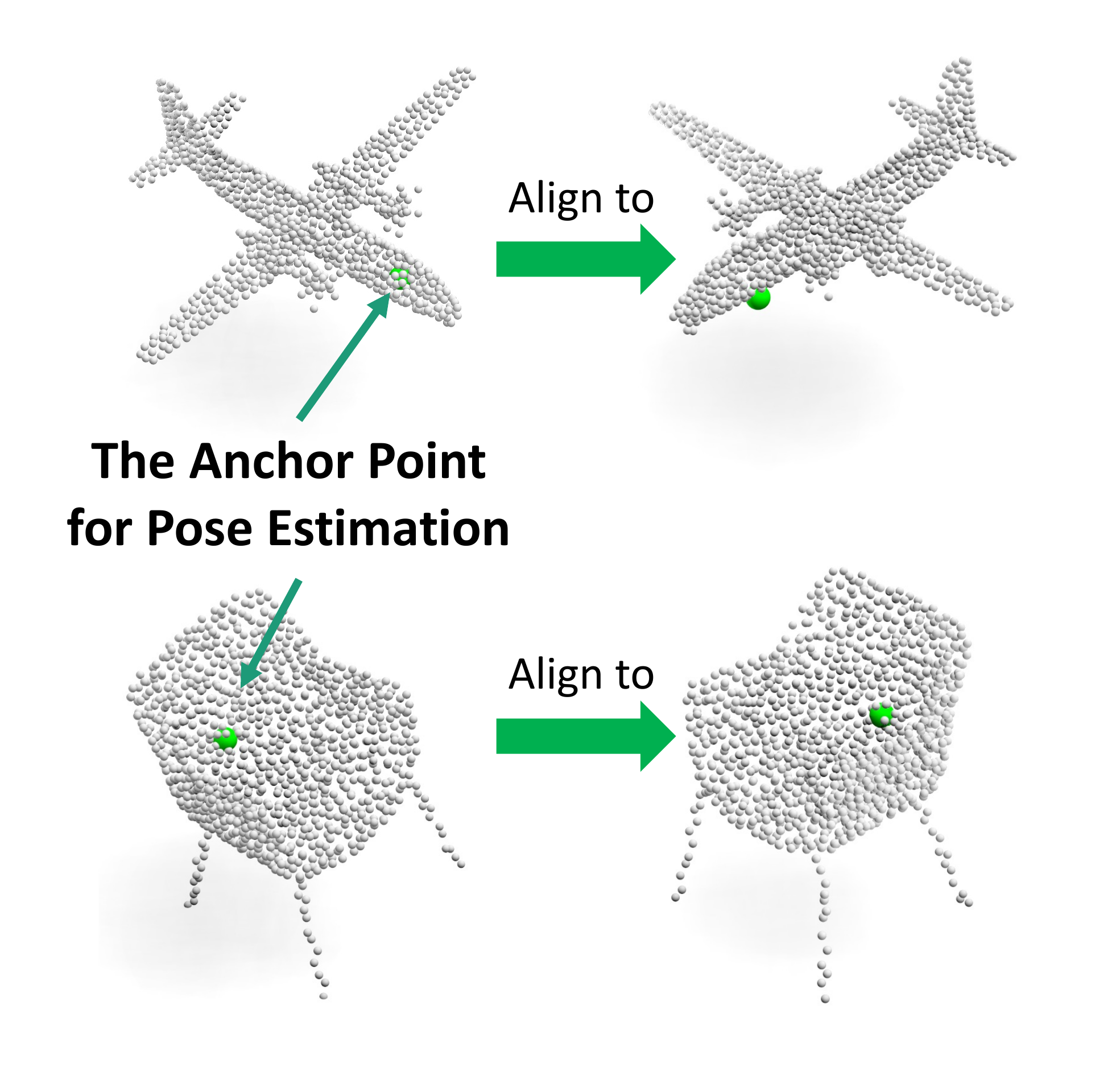}
    \end{center}
       \caption{An illustration of object alignment.}
    \label{fig:pose}
\end{figure}

As mentioned in the main paper, our CRIN can predict an anchor point for rotation estimation and further align the object, which is shown in Fig.~\ref{fig:pose}. Here we explain the procedure of the alignment. Given a randomly rotated point cloud as the source $\mathcal{P}_s$, and a target point cloud with the canonical pose denoted as $\mathcal{P}_t$. CRIN predicts an anchor point $p^*_s$ with its CRF $B^*_s$ for the source and  $p^*_t$ with $B^*_t$ for the target. The predicted rotation is:
\begin{equation}
    R^* = {B^*_t} {B^*_s}^T.
\end{equation}
Thus, the alignment procedure is formulated as:
\begin{equation}
\begin{split}
    \hat{\mathcal{P}}_s &= R^*  \mathcal{P}_s  \\
    &= {B^*_t} {B^*_s}^T \mathcal{P}_s,
\end{split}
\end{equation}
where $\hat{\mathcal{P}}_s$ represents the aligned source point cloud. Specifically, we first transform the source object to a rotation-invariant representation $B^*_s$ decided by the $p^*_s$. Then we use the basis $B^*_t$ of target point $p^*_t$ to transform it back to the target space. The key to aligning the object accurately to the target pose is finding the anchor point robust to the rotation transformations. Then we could localize the specific centrifugal axis which $B^*_s$ and $B^*_t$ share. In other words, we find the same rotation-invariant representation under different rotations. We have
\begin{equation}
    \begin{split}
        {B^*_s}^T  \mathcal{P}_s \approx  {B^*_t}^T \mathcal{P}_t
    \end{split}
\end{equation}
Therefore, CRIN can sample a stable anchor point invariant to rotations for rotation estimation.